\RequirePackage{fix-cm}
\documentclass[smallextended]{svjour3}       
\smartqed  
\usepackage{graphicx}
\usepackage{xcolor}
\usepackage{booktabs}       
\usepackage{hyperref}
\usepackage{caption}
\usepackage{subcaption}
\usepackage{amsmath}
\begin{document}

\title{FIRe-GAN: A novel Deep Learning-based infrared-visible fusion method for wildfire imagery \thanks{This research is supported in part by Tecnológico de Monterrey and the Mexican National Council of Science and Technology (CONACYT). This research is part of the project 7817-2019 funded by the Jalisco State Council of Science and Technology (COECYTJAL).}
}

\titlerunning{FIRe-GAN: A Deep Learning-based fusion method for wildfire imagery}        

\author{J. F. Ciprián-Sánchez         \and
        G. Ochoa-Ruiz \and M. Gonzalez-Mendoza \and
        L. Rossi 
}


\institute{J. F. Ciprián-Sánchez - Corresponding author\at
              Tecnológico de Monterrey,
              School of Engineering and Science,
              Av. Lago de Guadalupe KM 3.5, Margarita Maza de Juárez, 52926 Cd López Mateos, Mexico\\
              \email{A01373326@itesm.mx}           
           \and
           G. Ochoa-Ruiz and Miguel Gonzalez Mendoza \at
              Tecnológico de Monterrey, School of Engineering and Sciences, Av. Eugenio Garza Sada 2501, Monterrey, N.L., México 64849, Mexico \\
              \email{gilberto.ochoa@tec.mx}           
          \and
          L. Rossi \at
            Università di Corsica, Laboratoire Sciences Pour l’Environnement, Campus Grimaldi – BP 52 – 20250, Corti, France \\
            \email{rossi\_l@univ-corse.fr}
}

\date{Received: date / Accepted: date}

\maketitle

\begin{abstract}
Early wildfire detection is of paramount importance to avoid as much damage as possible to the environment, properties, and lives. Deep Learning (DL) models that can leverage both visible and infrared information have the potential to display state-of-the-art performance, with lower false-positive rates than existing techniques. However, most DL-based image fusion methods have not been evaluated in the domain of fire imagery. Additionally, to the best of our knowledge, no publicly available dataset contains visible-infrared fused fire images. There is a growing interest in DL-based image fusion techniques due to their reduced complexity. Due to the latter, we select three state-of-the-art, DL-based image fusion techniques and evaluate them for the specific task of fire image fusion. We compare the performance of these methods on selected metrics. Finally, we also present an extension to one of the said methods, that we called \emph{FIRe-GAN}, that improves the generation of artificial infrared images and fused ones on selected metrics. 
\end{abstract}

\section{Declarations}
\label{declarations}

\subsection{Funding}
\label{declarations:funding}
This research is supported in part by Tecnologico de Monterrey and the Mexican National Council of Science and Technology (CONACYT). This research is part of the project 7817-2019 funded by the Jalisco State Council of Science and Technology (COECYTJAL).


\subsection{Availability of data and material}
\label{declarations:availability_data}
The image fusion methods by Li et al. \cite{Li18} and Ma et al. \cite{Ma19_GAN} are publicly available as Github repositories in \cite{Li18_github,Ma19_GAN_github} respectively. The RGB-NIR dataset employed for pre-training the method by Zhao et al. ~\cite{Zhao20} was developed by Brown et al. ~\cite{Brown11} and is publicly available at ~\cite{RGB-NIRDataset}. The Corsican Fire Database \cite{Toulouse17}  is available upon request to the University of Corsica at ~\cite{CorsicanFireDatabase}.

\subsection{Code availability}
\label{declarations:code_availability}
The code generated by the authors implementing \emph{FIRe-GAN}, an extended version of the image fusion method proposed by Zhao et al. \cite{Zhao20}, is available as an open-source Github repository at \url{https://github.com/JorgeFCS/Image-fusion-GAN}.



\section{Introduction}
\label{intro}
Wildfires can occur naturally or due to human activities and have the potential to get out of control and have a significant impact on the environment, properties, and lives. Recently, there have been several wildfires of significant proportions worldwide, such as the Australian wildfires of 2019 and 2020. CNN reported that the said fires took the lives of at least 28 people ~\cite{CNN_Australia}. Another more recent example is the ongoing wildfire season in California in the US.  According to the BBC, as of September 17, 2020, 6.7 million acres have burned and more than 30 people have died ~\cite{BBC_California}. Early wildfire detection enabling technologies are thus crucial in order to avoid as much damage as possible and to help firefighters in their endeavors.

Vision-based fire detection techniques can be divided into visible or infrared fire detection systems ~\cite{Yuan15}, according to the spectral range of the cameras employed. In operative scenarios, visible image-based systems display significant false alarm rates and missed detections. The latter is due to constraints present in different situations, such as changing environmental conditions and illumination ~\cite{Cetin13}. In contrast, infrared cameras can perform flame detection in weak or no light conditions. Additionally, smoke is transparent in these kinds of images. As such, it can be practical to employ infrared-based systems to perform flame detection in both daytime and nighttime conditions ~\cite{Yuan15}. There has been work on fire detection on the NIR and LWIR infrared bands; however, it is also not trivial to detect fire on infrared images, as they present problems such as thermal reflections and IR blocking ~\cite{Cetin13}.

The fusion of visible and infrared images can be beneficial for improving the robustness, accuracy, and reliability of fire detection systems ~\cite{Yuan15}. Although there have been some approaches in this area ~\cite{Arrue2000,Martinez-de-Dios08}, as well as for fire segmentation ~\cite{Nemalidinne18}, to the best of our knowledge, DL-based visible-infrared fusion methods have not been tested for the particular task of fire image fusion.

There is a growing interest in DL-based image fusion techniques. The latter is due to their reduced complexity compared to methods on the multi-scale transform and representation learning domains ~\cite{Li20}. In order to assess the applicability of some of the most promising DL-based approaches in IR-visible fusion to wildfire image fusion, we chose three state-of-the-art methods and evaluate them on the particular task of fire image fusion. We also implement extensions on one of them, generating the proposed ~\emph{FIRe-GAN} method to improve its applicability to the Corsican Dataset, introduced in this paper.

The selected methods are as follows: the method proposed by Li et al. ~\cite{Li18}, the work by Ma et al. ~\cite{Ma19_GAN}, and the architecture proposed by Zhao et al. ~\cite{Zhao20}. We evaluate and compare these methods on the Corsican Fire Database ~\cite{Toulouse17} using some of the most important metrics for assessing image quality in the literature.

These state of the art methods were selected because they present several desirable features. The method by Li et al. ~\cite{Li18} uses a pre-trained VGG19 Deep Convolutional Neural Network (DCNN) as a part of its process for the fusion of detail content. Since the authors employ only selected layers of the network, no further training on new datasets (such as ours) is needed. 

The method by Ma et al. ~\cite{Ma19_GAN} represents, to the best of our knowledge, the first approach towards image fusion through the use of Generative Adversarial Networks (GANs). This technique has the advantage of being end-to-end, which significantly reduces its implementation and training complexity. 

The method by Zhao et al. ~\cite{Zhao20} is a GAN-based approach as well, with the additional feature of being able to generate approximate infrared images from visible ones. It is relevant to note that the type of infrared images (near-infrared (NIR), short-wavelength (SWIR), mid-wavelength (MWIR), or long-wavelength (LWIR)) that the model learns to generate depends on the training set. It would learn to generate NIR images if those are the ones contained in the training set, and so forth. 


Finally, it is relevant to note that many visible-infrared fusion methods ~\cite{Li18,Ma19_GAN,Zhao20,Li20,Zhao20_decomposition,Ma20_adversarial} output grayscale fused images, which means that the color information of the visible image is lost. In the present paper, we present the ~\emph{FIRe-GAN} model, an extended version of the method proposed by Zhao et al. ~\cite{Zhao20}, which allows for the processing of higher resolution images and the generation of color fused images as outputs. The latter is relevant due to color being one of the most used features in visible image-based fire detection methods ~\cite{Yuan15}.

The main contributions of this article are as follows:
~\begin{enumerate}
    \item We provide a thorough analysis of existing DL-fusion methods for conventional imagery.
    \item We provide a quantitative demonstration of the feasibility of applying DL-based fusion methods for infrared imagery from wildfires.
    \item We introduce a novel artificial IR and fused image generator that has been tested both in conventional and fire imagery.
\end{enumerate}

We believe that these contributions can potentially boost developments in the wildfire fighting community that makes use of visible and infrared fire imagery to perform accurate and reliable wildfire detection and monitoring. It must be noted that this work is part of a larger endeavour in which the proposed fusion methods plays only a small but vital role, and the generation of infrared images, which is a component of the system, is to be used in other related research efforts and could prove to be useful for the research community at large. 






The rest of this paper proceeds as follows. In Section ~\ref{methods_data} present the three compared methods, the evaluation metrics, and the datasets employed. Section ~\ref{results} describes the experiments and shows the evaluation results and a quantitative comparison of the selected methods. In Section ~\ref{discussion} we provide a discussion on the obtained results. Finally, section ~\ref{conclusions_future_work} presents the conclusions and outlines potential future work.

\section{Methods and data}
\label{methods_data}

\subsection{Evaluated methods}
\label{evaluated_methods}

\subsubsection{Infrared and visible image fusion using a deep learning framework}
\label{explanation_li}

This method, proposed by Li et al. ~\cite{Li18}  employs a DL framework for the generation of a single image that contains all the features present in visible and infrared images. 

First,  the original images are decomposed into base parts and detail content. Then, these base parts are fused through weight-averaging. For the detail parts, the authors employ a DCNN network to extract multi-layer features ~\cite{Li18_github}. With the extracted features, L1 normalization coupled with a weighted average strategy is employed in order to generate candidates for the fused detail content. Afterwards, a max selection strategy is used to obtain the final fused detail content. Finally, the final fused image is constructed by combining the fused detail and base contents. 

It is worth noting that the authors employ the fixed VGG19 network pre-trained in ImageNet, for the extraction of the multi-layer features. We will be referring to this method as the $VGG19$ method.


\subsubsection{FusionGAN: A generative adversarial network for infrared and visible image fusion}

This method, proposed by Ma et al. introduces an image fusion method based on a Generative Adversarial Network (GAN). This work is, to the best of our knowledge, the first time a GAN model is proposed for the image fusion task. The architecture is an end-to-end model that generates the fused image automatically from the source images with no need to define fusion rules.

The generator attempts to produce a fused image with significant thermal information, as well as with gradients from the visible image. The discriminator, in turn, forces the generated image to contain more details from the visible image. The latter module enables the model to produce images that retain both thermal and textural information. Finally, the authors generalize their proposed model so that it can fuse images with different resolutions, with the final image free of the noise caused by the upsampling of infrared information. Ma et al. named this model as the $FusionGAN$ model.


\subsubsection{Fusion of Unmatched Infrared and Visible Images Based on Generative Adversarial Networks}

This method was proposed by Zhao et al. ~\cite{Zhao20}. In this work, the authors propose a network model based on generative adversarial networks (GANs) to fuse unmatched infrared and visible images. First, the visible image is given as an input to the generator $G_1$ to create a synthetic infrared image. Then, the visible image and the synthetic infrared image are concatenated and input into generator $G_2$, generating the fused image as the output. The discriminator $D_1$ distinguishes between the real visible image and the generated fused image so that the fused image is closer to the visible image, containing more textural details. Simultaneously, the discriminator $D_2$ distinguishes the real infrared image, the generated infrared image, and the fused image. Through the updating cycle, the generated infrared image becomes closer to the real infrared image, allowing the fused image to contain more thermal information. We will be referring to this model as the $UnmatchGAN$ model.

\subsubsection{Summary}
\label{evaluated_methods_summary}

The previously mentioned methods have both advantages and disadvantages. The framework proposed by Li et al. ~\cite{Li18} has the advantage of only needing some layers of an already pre-trained VGG19 network to perform feature extraction. As such, there is no need for further training for the particular application of image fusion. However, it is not an end-to-end method, and the required intermediate steps increase its implementation complexity.

The model presented by Ma et al. ~\cite{Ma19_GAN} has the advantage of being an end-to-end model, significantly reducing its implementation complexity. However, this GAN-based method needs to train on visible-infrared image pairs, and in consequence, its performance depends on the quality of the training process. It is also relevant to note that GANs have the challenge of training stability ~\cite{Miyato18}.

The model proposed by Zhao et al. ~\cite{Zhao20}, a GAN-based model too, has the advantage of being an end-to-end procedure; however, the challenge lies on the training stability, as well as the need for a good training dataset. This method has the additional capability of learning to generate approximate infrared images based on source visible ones. Since the fusion process requires perfectly aligned source images ~\cite{Ma19_survey}, for the particular context of fire images, this could prove a significant advantage for the research community given the burden of obtaining perfectly matched visible-infrared fire images on realistic operative scenarios.

Finally, it is also relevant to note that the three methods output grayscale fused images. In the context of fire imagery, the preservation of color could prove beneficial.

\subsection{Data}
\label{data}

For the present paper, we employ the visible-infrared image pairs of the Corsican Fire Dataset, first presented by Toulouse et al. ~\cite{Toulouse17} in the \emph{Fire Safety Journal} in 2017. This dataset contains 640 pairs of visible and infrared fire images, alongside their corresponding ground truths for fire region segmentation.

\begin{figure*}[htb]
    \centering 

\begin{subfigure}{0.23\textwidth}
  \includegraphics[width=\linewidth]{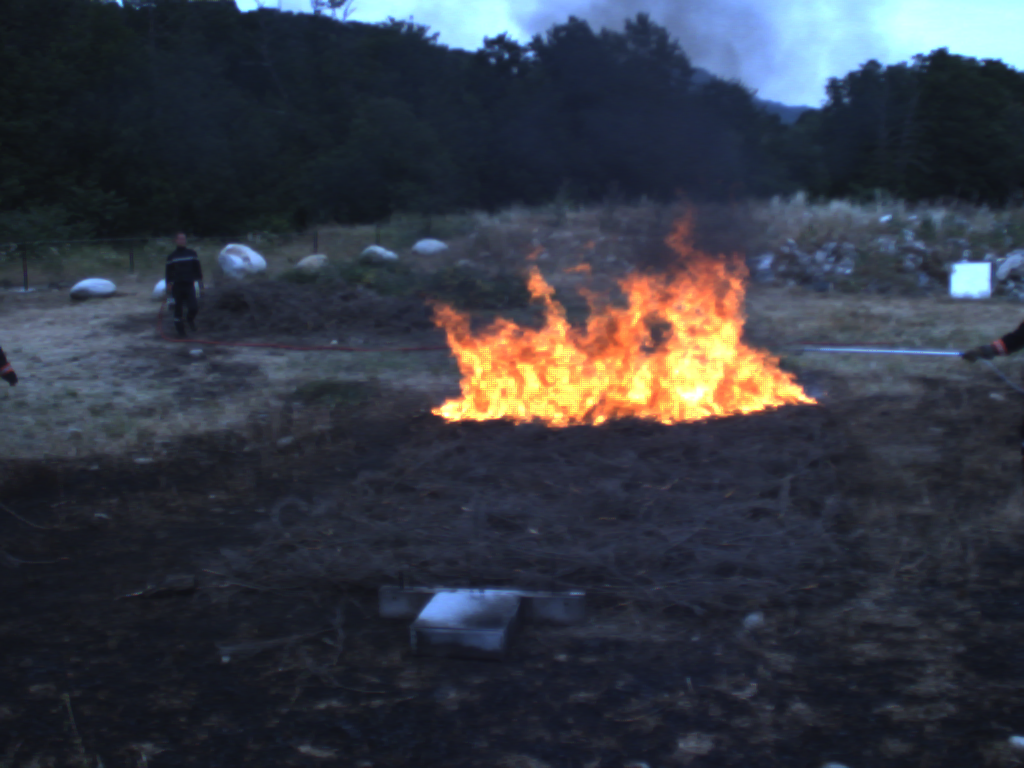}
  \caption{Fire - visible}
  \label{fig:rgb_samples_fire}
\end{subfigure}\hfil 
\begin{subfigure}{0.23\textwidth}
  \includegraphics[width=\linewidth]{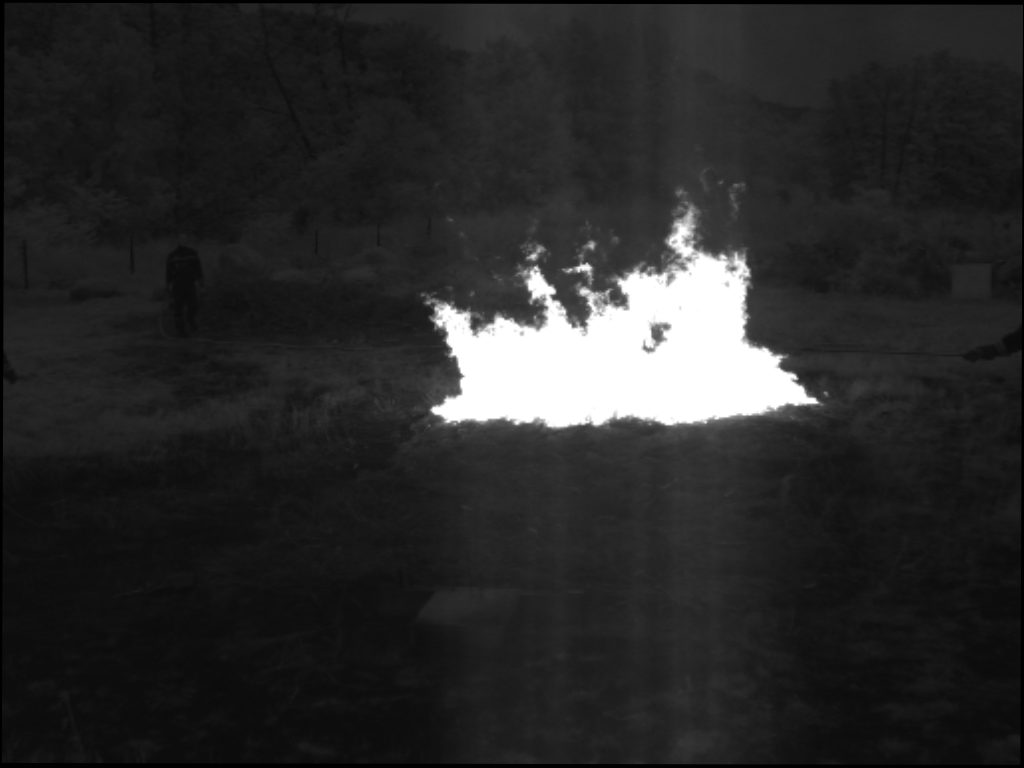}
  \caption{Fire - NIR}
  \label{fig:ir_samples_fire}
\end{subfigure}\hfil 
\begin{subfigure}{0.23\textwidth}
  \includegraphics[width=\linewidth]{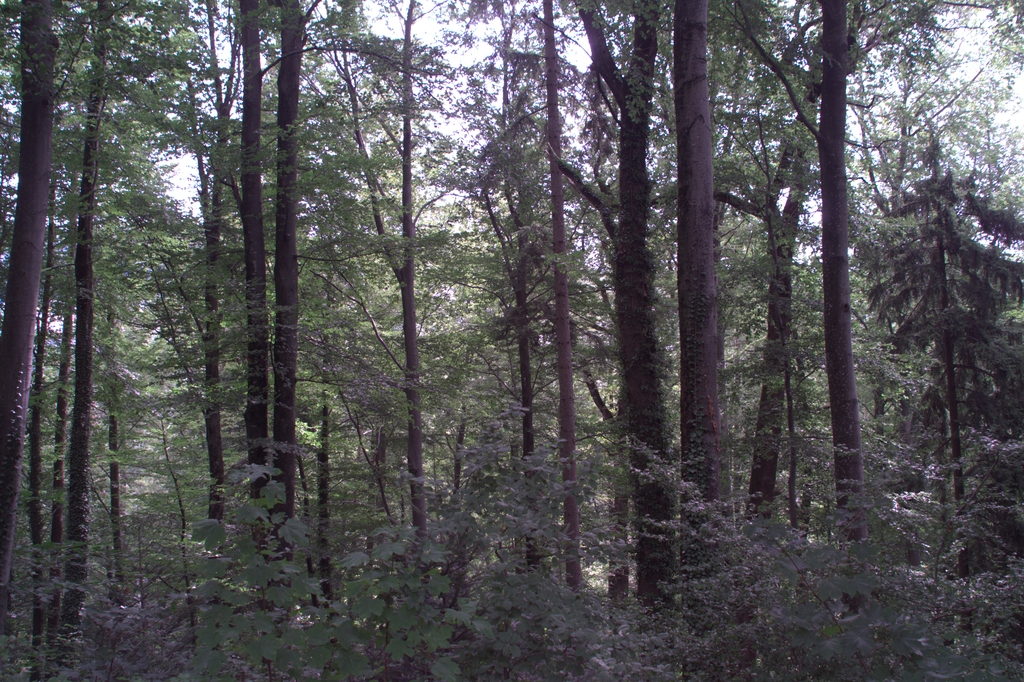}
  \caption{Non-fire - visible}
  \label{fig:rgb_samples_no_fire}
\end{subfigure}\hfil 
\begin{subfigure}{0.23\textwidth}
  \includegraphics[width=\linewidth]{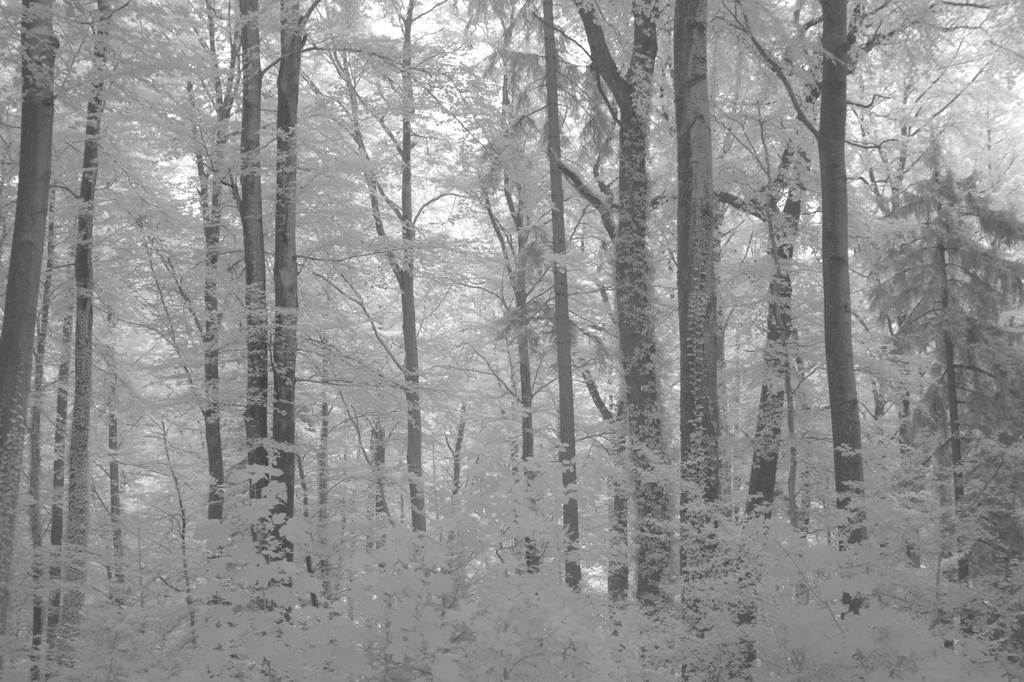}
  \caption{Non-fire - NIR}
  \label{fig:ir_samples_no_fire}
\end{subfigure}\hfil 

\caption{Sample images for the RGB-NIR and Corsican Fire Database datasets.}
\label{fig:dataset_samples}
\end{figure*}

We also employ the RGB-NIR dataset, developed by Brown et al. ~\cite{Brown11} and presented at \emph{CVPR 2011}. This dataset contains 477 non-fire visible-infrared image pairs. Fig. ~\ref{fig:dataset_samples} shows sample images from both datasets. It is relevant to note that the infrared images are NIR ones. 

\subsection{Metrics}
\label{metrics}

The metrics selected for the present paper are the information entropy (EN), the correlation coefficient (CC), the peak signal-to-noise-ratio (PSNR), and the structural similarity index measure (SSIM); these metrics are by far the more common in this area, more details can be found in \cite{metrics}. In the following subsections, we describe succinctly these metrics.

\subsubsection{Information entropy}
EN reflects the average amount of information in an image. It is defined as: 

\begin{equation}
EN = - \sum_{l=1}^{L-1} p_l \log_2 p_l,
\end{equation}

where $L$ stands for the gray levels of the image, and $p_l$ represents the proportion of gray-valued pixels $i$ in the total number of pixels. The larger EN is, the more information is in the fused image ~\cite{Zhao20}.

\subsubsection{Correlation coefficient}
The CC measures the degree of linear correlation between the fused image and either the visible or infrared image. It is defined as: 

\begin{equation}
CC(X,Y) = \frac{Cov(X,Y)}{\sqrt{Var(X) Var(Y)}},
\end{equation}

where $Cov(X,Y)$ is the covariance between the fused image and the reference images, and $Var(X)$, $Var(Y)$ represent the variance of the two images. The larger the value of CC, the higher the correlation between the fused and the reference images ~\cite{Zhao20}.

\subsubsection{Peak signal-to-noise ratio}
The PSNR assumes that the difference between the fused image and the reference image is noise. It is defined as:

\begin{equation}
PSNR = 10\log_{10}(\frac{MAX^2}{MSE}),
\end{equation}

where $MAX$ is the maximum value of the image color, and $MSE$ is the mean squared error. An accepted benchmark for this metric is 30 dB; a PSNR value lower than this threshold means that the fused image presents significant deterioration ~\cite{Zhao20}.

\subsubsection{Structural similarity index measure}
The SSIM is a method for measuring the similarity between two images ~\cite{Wang11}. It is based on the degradation of structural information ~\cite{Wang04} and is defined as follows:

\begin{equation}
SSIM(X,Y) = (\frac{2 u_x u_y + c_1}{u_x^2 + u_y^2 + c_1 })^\alpha * (\frac{2 \sigma_x \sigma_y + c_2}{\sigma_x^2 + \sigma_y^2 + c_2 })^\beta * \\(\frac{\sigma_{xy} + c_3}{\sigma_x \sigma_y + c_3 })^\gamma,
\end{equation}

where $x$ and $y$ are the reference and fused images, respectively; $u_x$, $u_y$, $\sigma_x^2$, $\sigma_y^2$, and $\sigma_{xy}$ represent the mean value, variance, and covariance of images $x$ and $y$, respectively. Finally, $c_1$, $c_2$, and $c_3$ are small numbers that help to avoid a division by zero, and $\alpha$, $\beta$, and $\gamma$ are used to adjust the proportions \cite{Zhao20}.

The range of values for SSIM goes from 0 to 1, with 1 being the best possible one.

\section{Results}
\label{results}

We evaluate the methods by Li et al. ~\cite{Li18} and Ma et al. ~\cite{Ma19_GAN} through their corresponding open-source implementations. These models are provided by the authors pre-trained on the ImageNet and the TNO Dataset, respectively.

\subsection{Architectural adjustment for wildfire imagery fusion}
\label{arch_modification}

Since the method proposed by Zhao et al. ~\cite{Zhao20} has no available open-source implementation, we implemented it from the ground-up, extending it into our proposed ~\emph{FIRe-GAN} model. We refer to the previous work by Isola et al. ~\cite{Isola17} for relevant implementation details on G1 and the work by Ma et al. ~\cite{Ma19_GAN} for G2 and both discriminators. We also modified the final layer of both generators from 1 to 3 filters; the latter allows our architecture to output 3-channel images. Since fire images are generally of a high size and resolution, we made use of a U-Net architecture for G1 in contrast to the original method that makes use of a simple enconder-decoder architecture. Additionally, we integrated the Two Time-Scale Update Rule (TTUR) module proposed by Heusel et al. ~\cite{Heusel17} and spectral normalization to the discriminators, as per the work by Miyato et al. ~\cite{Miyato18} to increase the training stability.

\begin{figure}
  \includegraphics[trim=25 420 0 140,clip,scale=0.4]{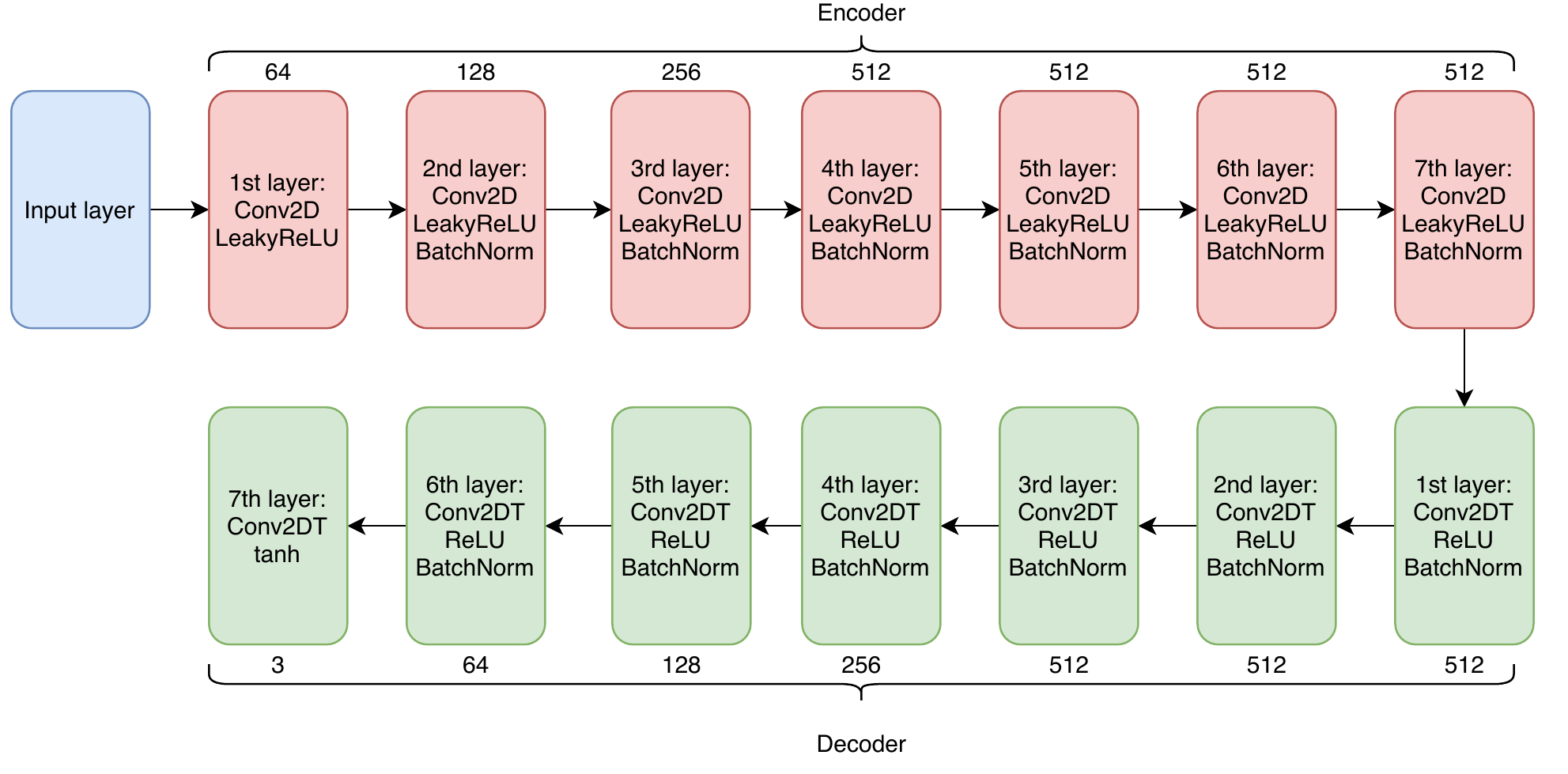}
\caption{Implemented architecture for G1 of the method by Zhao et al. ~\cite{Zhao20} with the mentioned considerations.}
\label{fig:G1}       
\end{figure}

\begin{figure}
  \includegraphics[trim=10 420 0 140,clip,scale=0.4]{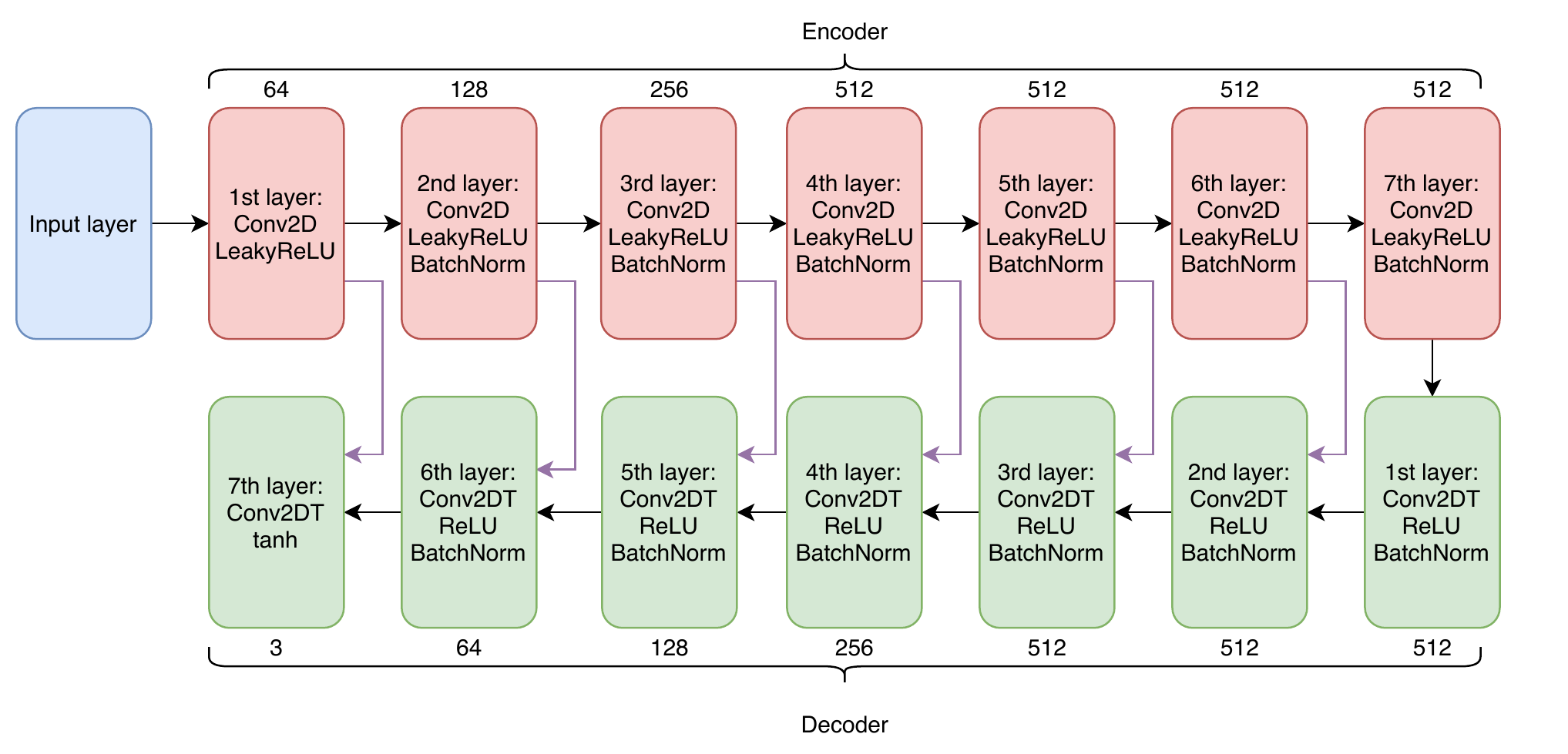}
\caption{Implemented architecture for G1 with the proposed U-Net architecture.}
\label{fig:G1_unet}       
\end{figure}

\begin{figure}
  \includegraphics[trim=25 250 155 475,clip,scale=0.81]{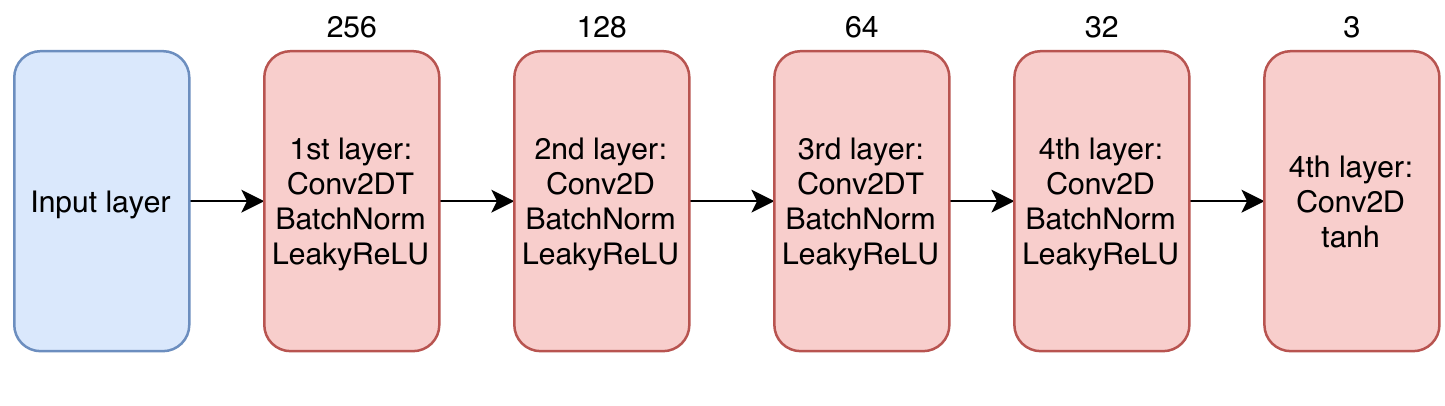}
\caption{Implemented architecture for G2 with the mentioned changes.}
\label{fig:G2}       
\end{figure}

\begin{figure}
  \includegraphics[trim=8 280 50 450,clip,scale=0.62]{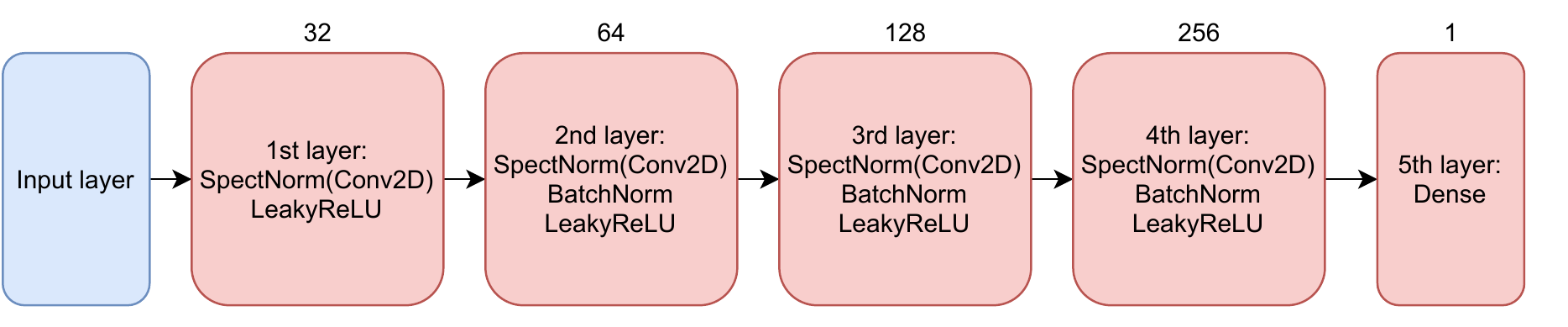}
\caption{Implemented architecture for both discriminators with the mentioned changes.}
\label{fig:Disc}       
\end{figure}

Fig. ~\ref{fig:G1} shows the architecture of G1 with the original enconder-decoder architecture, Fig. ~\ref{fig:G1_unet} presents G1 with the proposed U-Net architecture, Fig. ~\ref{fig:G2} shows the architecture of G2, and Fig. ~\ref{fig:Disc} the architecture of the discriminators, as per the said considerations.

To determine if the proposed U-Net architecture on G1 improves the quality of the generated infrared images, we pre-train the model with both architectures on the RGB-NIR dataset and compare the obtained results for the generated infrared images. We split the RGB-NIR dataset into a train and a validation sets. The training set contains 6112 image pairs after performing data augmentation, and the validation set consists of 96 image pairs. We trained the model with a batch size of 4, 40 epochs, a learning rate for both generators of 5e-5 and for both discriminators of 1e-4, and spectral normalization on both discriminators. Additionally, the discriminators were updated once every two generator updates.

\begin{figure*}[htb]
    \centering 
\begin{subfigure}{0.20\textwidth}
  \includegraphics[width=\linewidth]{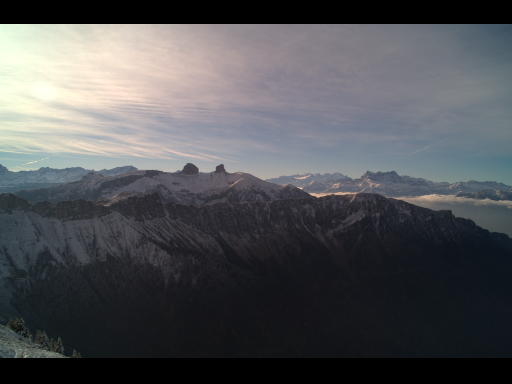}
  \caption{RGB images}
  \label{fig:unet_rgb}
\end{subfigure}\hfil 
\begin{subfigure}{0.20\textwidth}
  \includegraphics[width=\linewidth]{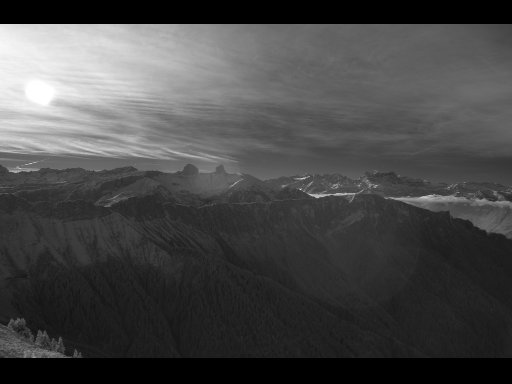}
  \caption{IR images}
  \label{fig:unet_ir}
\end{subfigure}\hfil 
\begin{subfigure}{0.20\textwidth}
  \includegraphics[width=\linewidth]{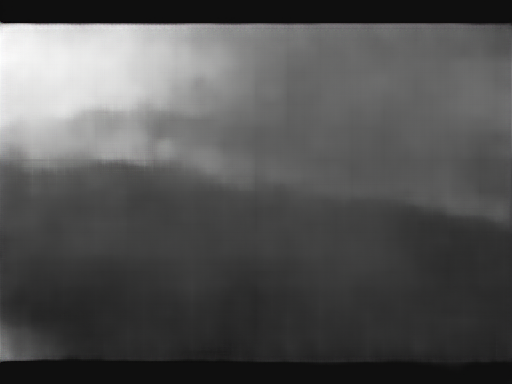}
  \caption{Original}
  \label{fig:unet_gir}
\end{subfigure}\hfil 
\begin{subfigure}{0.20\textwidth}
  \includegraphics[width=\linewidth]{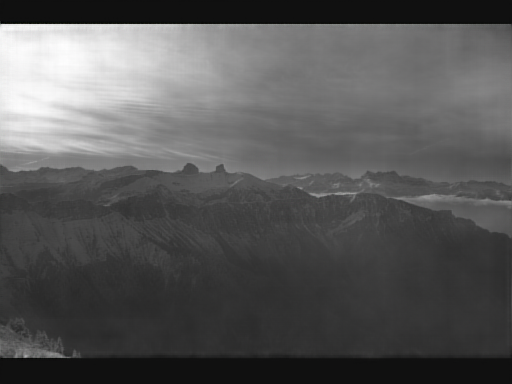}
  \caption{U-Net}
  \label{fig:unet_gir2}
\end{subfigure}
\caption{Sample resulting artificial infrared images from both architectures for G1.}
\label{fig:unet_comparison}
\end{figure*}

In Table ~\ref{tab:comp_unet_g1} we present the average results for both architectures in terms of the selected metrics. In this case, the CC, PSNR and SSIM metrics refer to the comparison of the source and generated infrared (IR) images. Fig. ~\ref{fig:unet_comparison} displays a sample of the images produced by both architectures. We can observe improvements on the CC, PSNR, and SSIM metrics for our proposed U-Net architecture. A visual assessment of the produced images allows us to note an increased amount of detail as well.


\begin{table}
\caption{Average results for both architectures of G1 on the generated IR images.}
\label{tab:comp_unet_g1}       
\begin{tabular}{lllll}
\hline\noalign{\smallskip}
Model      & EN      & CC       & PSNR    & SSIM   \\
\noalign{\smallskip}\hline\noalign{\smallskip}
Original   & 9.9158  & 0.8593   & 17.9341 & 0.5506 \\
U-Net      & 9.9474  & 0.9203   & 19.9473 & 0.7541 \\
\noalign{\smallskip}\hline
\end{tabular}
\end{table}

\subsection{Method comparison}
\label{method_comp}

\begin{figure*}
     \centering
     \begin{subfigure}[b]{0.43\textwidth}
         \centering
         \includegraphics[width=\textwidth]{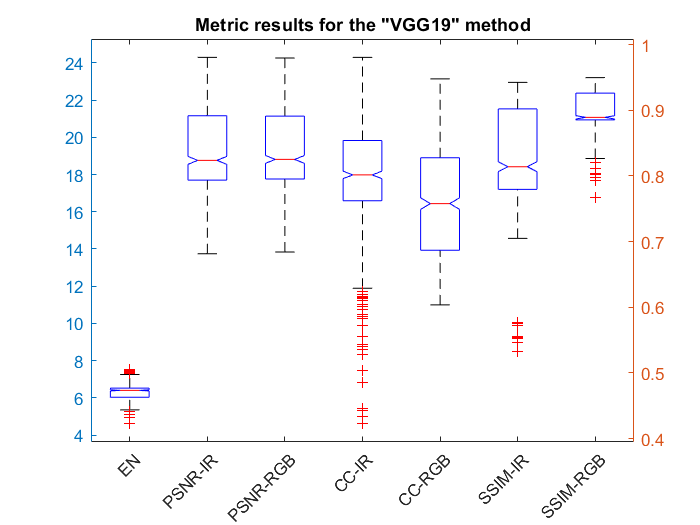}
         \caption{\emph{VGG19} method.}
         \label{fig:vgg19_fire}
     \end{subfigure}
     \hfill
     \begin{subfigure}[b]{0.43\textwidth}
         \centering
         \includegraphics[width=\textwidth]{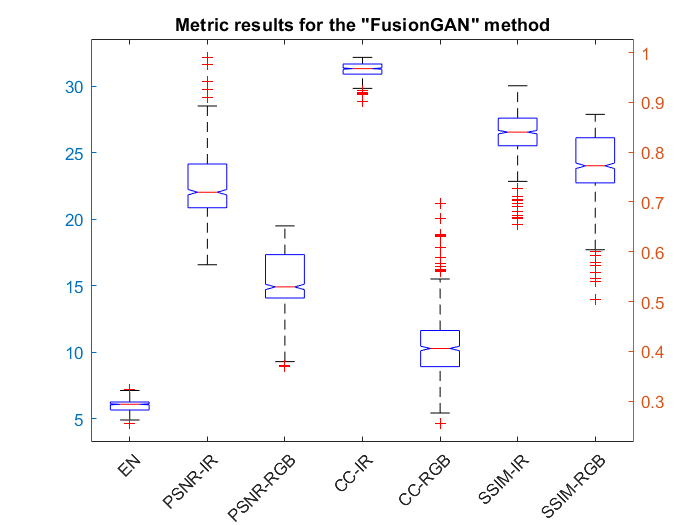}
         \caption{\emph{FusionGAN} method.}
         \label{fig:fusiongan_fire}
     \end{subfigure}
        
    \medskip
    
    \begin{subfigure}[b]{0.43\textwidth}
         \centering
         \includegraphics[width=\textwidth]{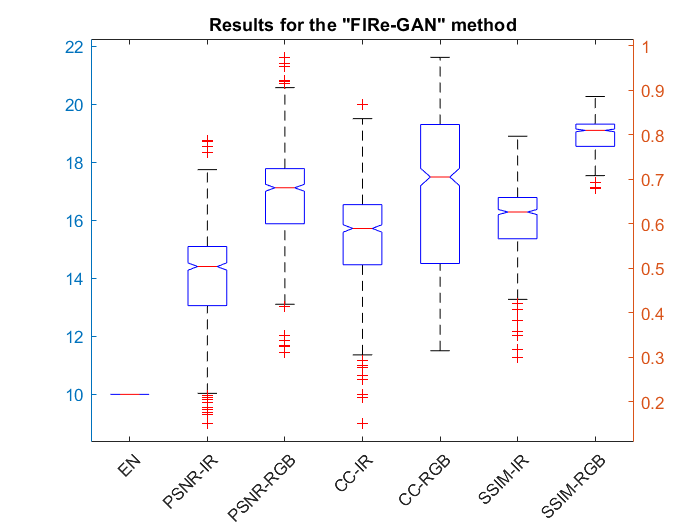}
         \caption{Proposed \emph{FIRe-GAN} method.}
         \label{fig:firegan_fire}
     \end{subfigure}
        \caption{Results from the fire images on the three evaluated methods.}
        \label{fig:three_methods_boxplots}
\end{figure*}

\begin{figure*}[htb]
    \centering 

\begin{subfigure}{0.19\textwidth}
  \includegraphics[width=\linewidth]{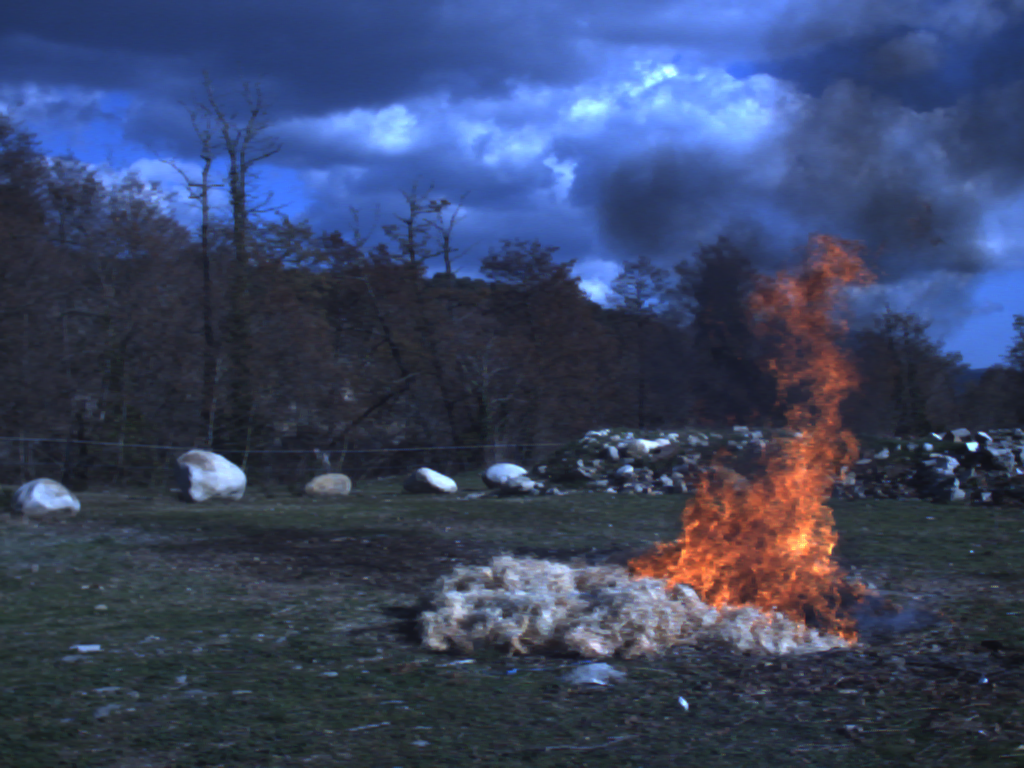}
\end{subfigure}\hfil 
\begin{subfigure}{0.19\textwidth}
  \includegraphics[width=\linewidth]{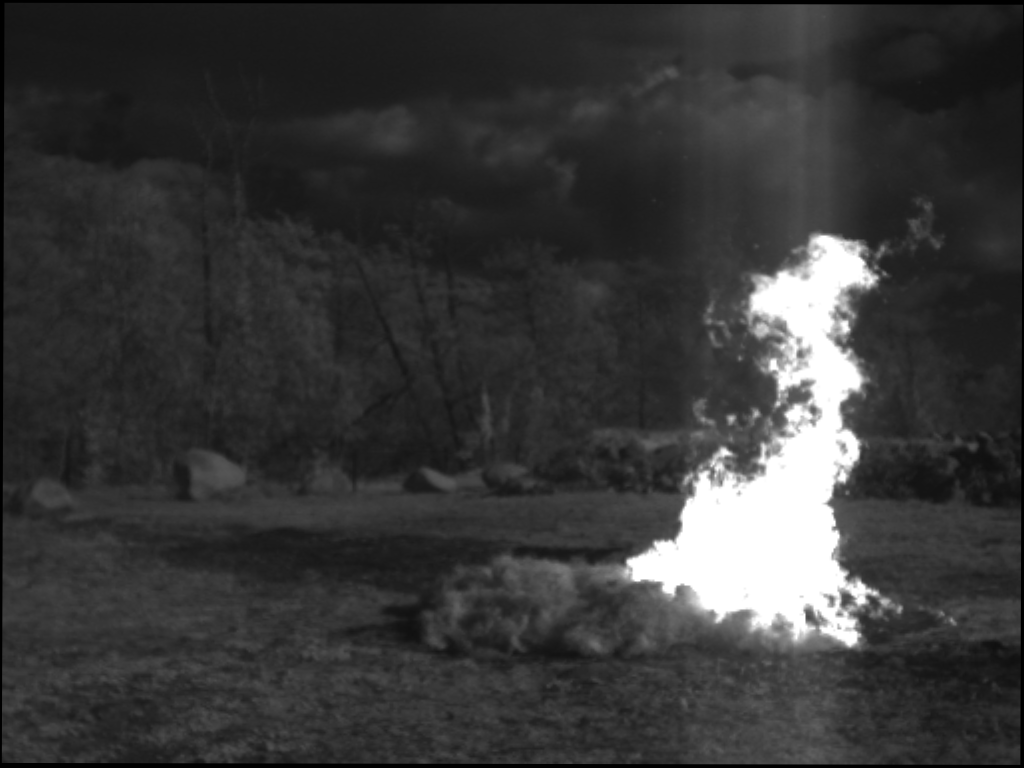}
\end{subfigure}\hfil 
\begin{subfigure}{0.19\textwidth}
  \includegraphics[width=\linewidth]{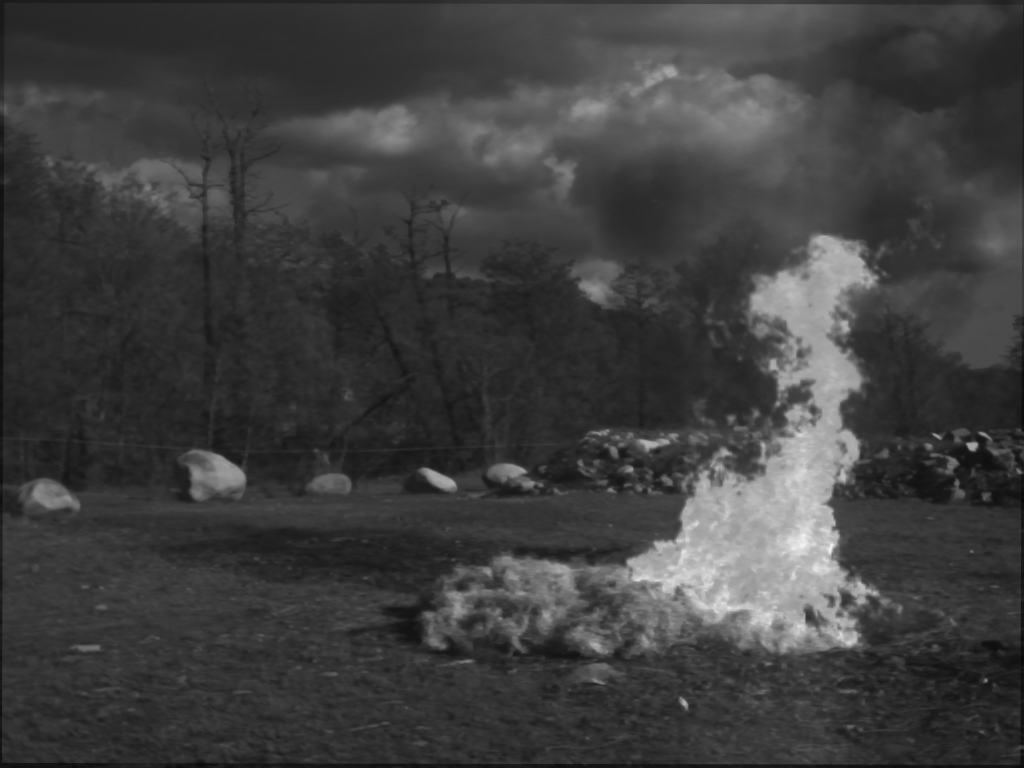}
\end{subfigure}\hfil 
\begin{subfigure}{0.19\textwidth}
  \includegraphics[width=\linewidth]{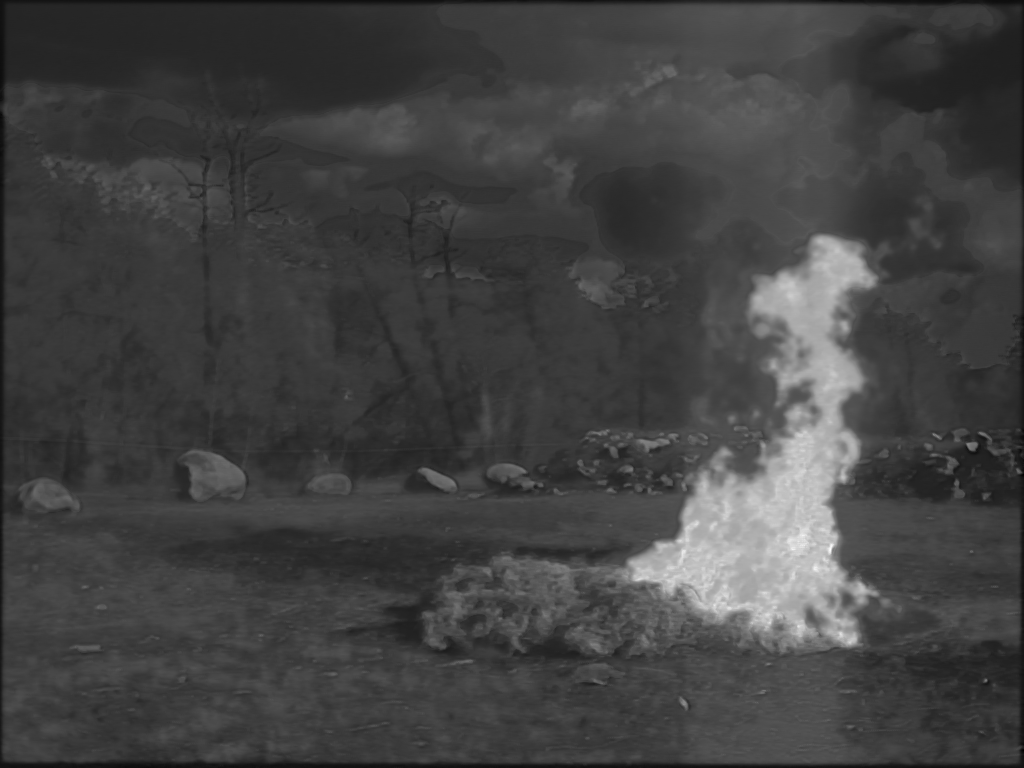}
\end{subfigure}\hfil 
\begin{subfigure}{0.19\textwidth}
  \includegraphics[width=\linewidth]{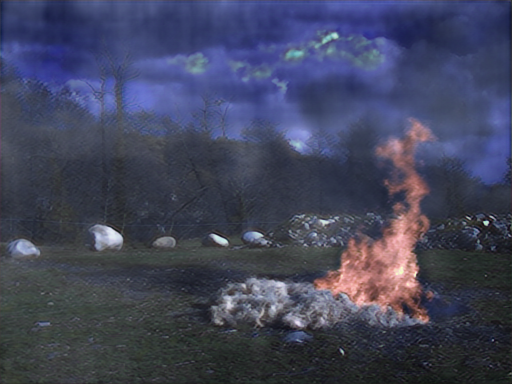}
\end{subfigure}

\medskip
\begin{subfigure}{0.19\textwidth}
  \includegraphics[width=\linewidth]{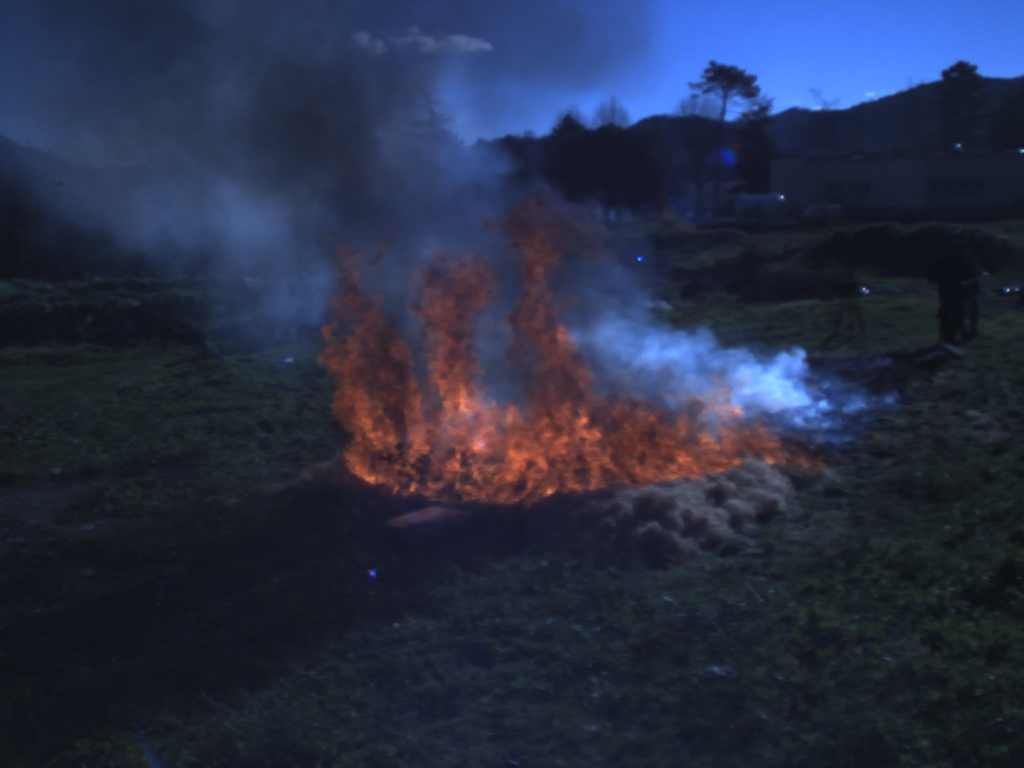}
\end{subfigure}\hfil 
\begin{subfigure}{0.19\textwidth}
  \includegraphics[width=\linewidth]{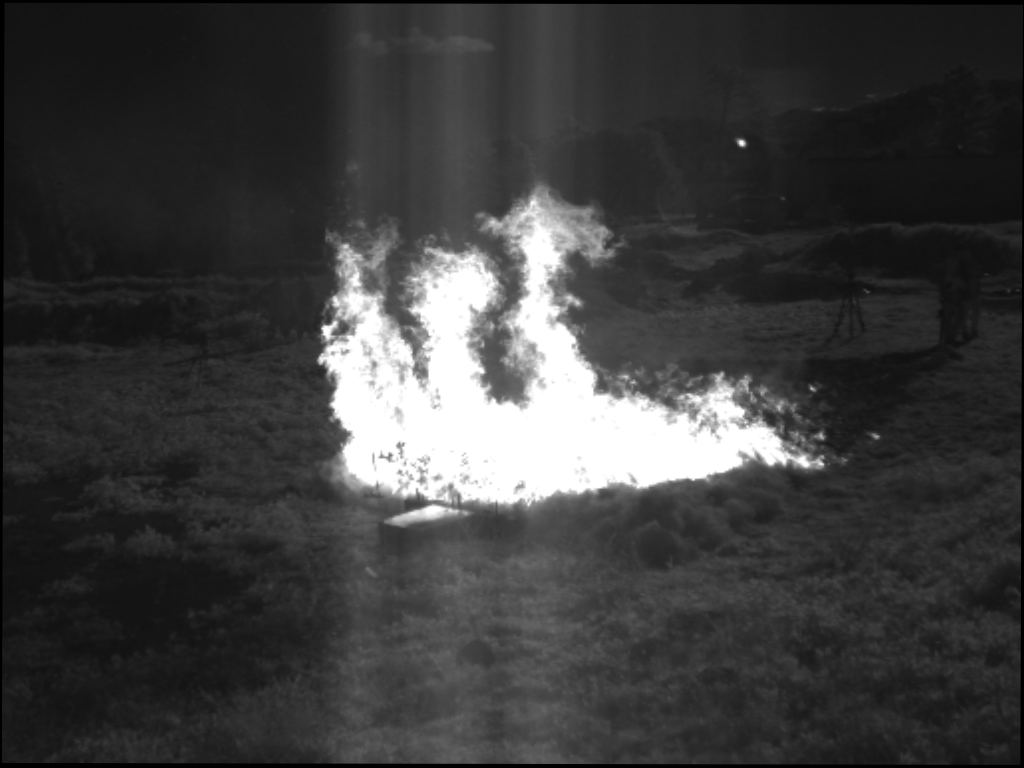}
\end{subfigure}\hfil 
\begin{subfigure}{0.19\textwidth}
  \includegraphics[width=\linewidth]{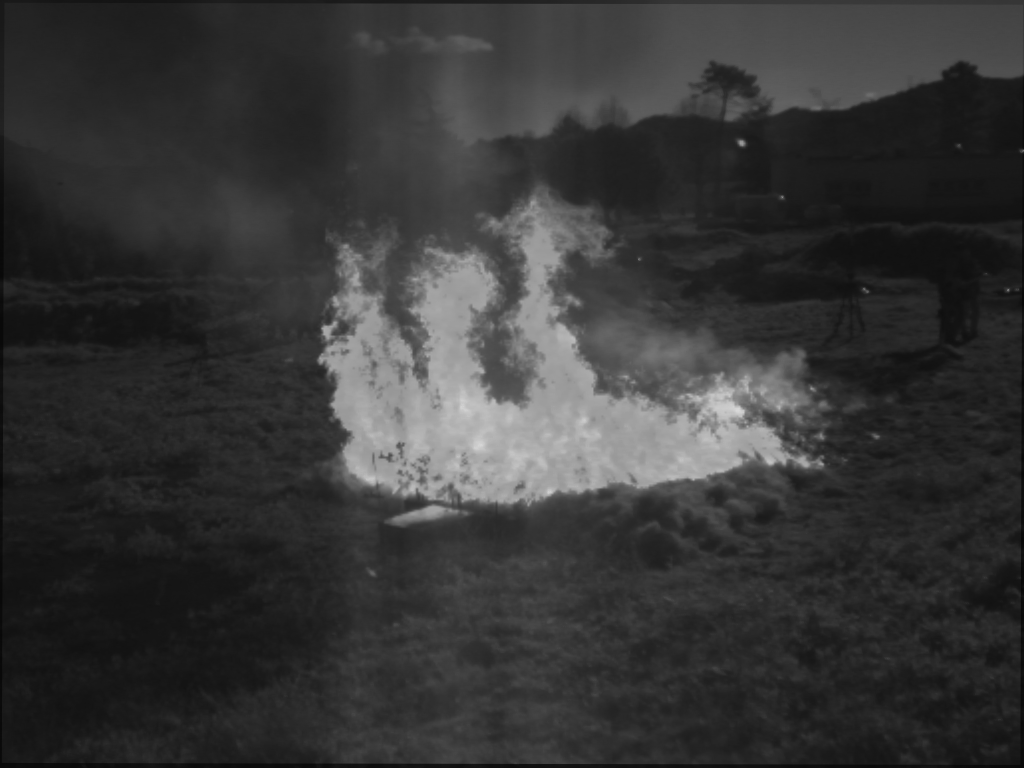}
\end{subfigure}\hfil 
\begin{subfigure}{0.19\textwidth}
  \includegraphics[width=\linewidth]{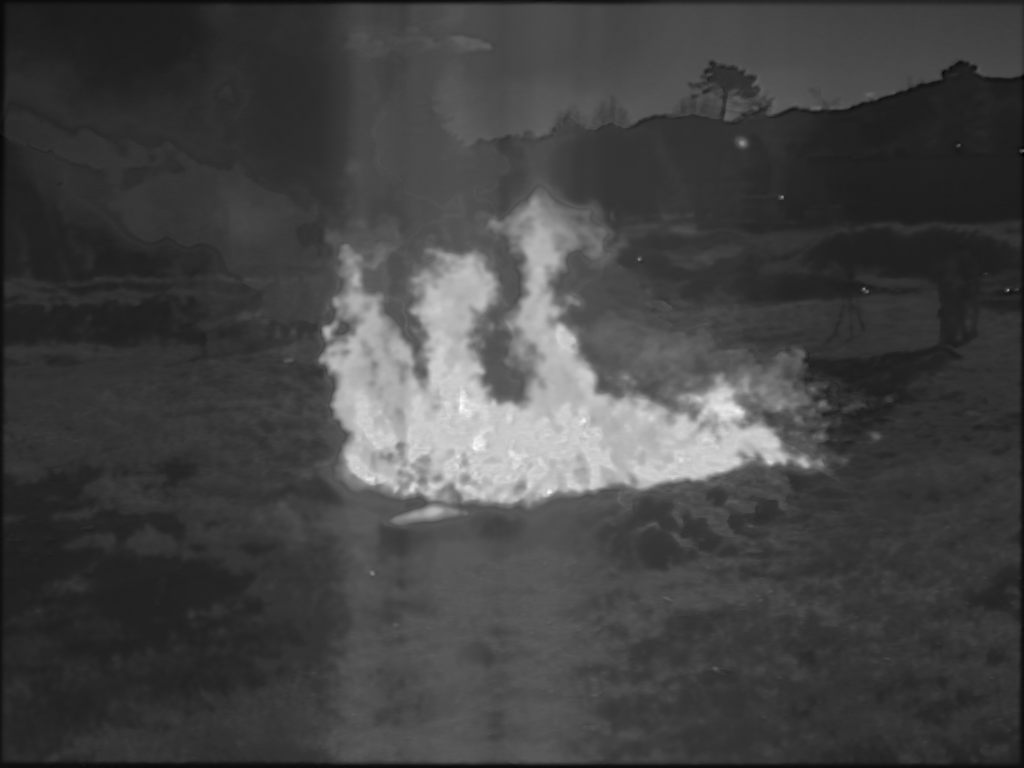}
\end{subfigure}\hfil 
\begin{subfigure}{0.19\textwidth}
  \includegraphics[width=\linewidth]{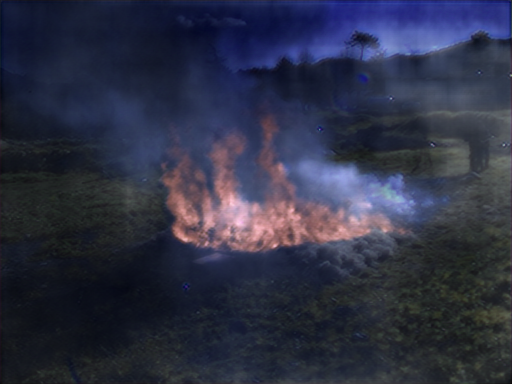}
\end{subfigure}

\medskip
\begin{subfigure}{0.19\textwidth}
  \includegraphics[width=\linewidth]{078_rgb.png}
  \caption{}
  \label{fig:rgb_three_methods}
\end{subfigure}\hfil 
\begin{subfigure}{0.19\textwidth}
  \includegraphics[width=\linewidth]{078_nir.png}
  \caption{}
  \label{fig:ir_three_methods}
\end{subfigure}\hfil 
\begin{subfigure}{0.19\textwidth}
  \includegraphics[width=\linewidth]{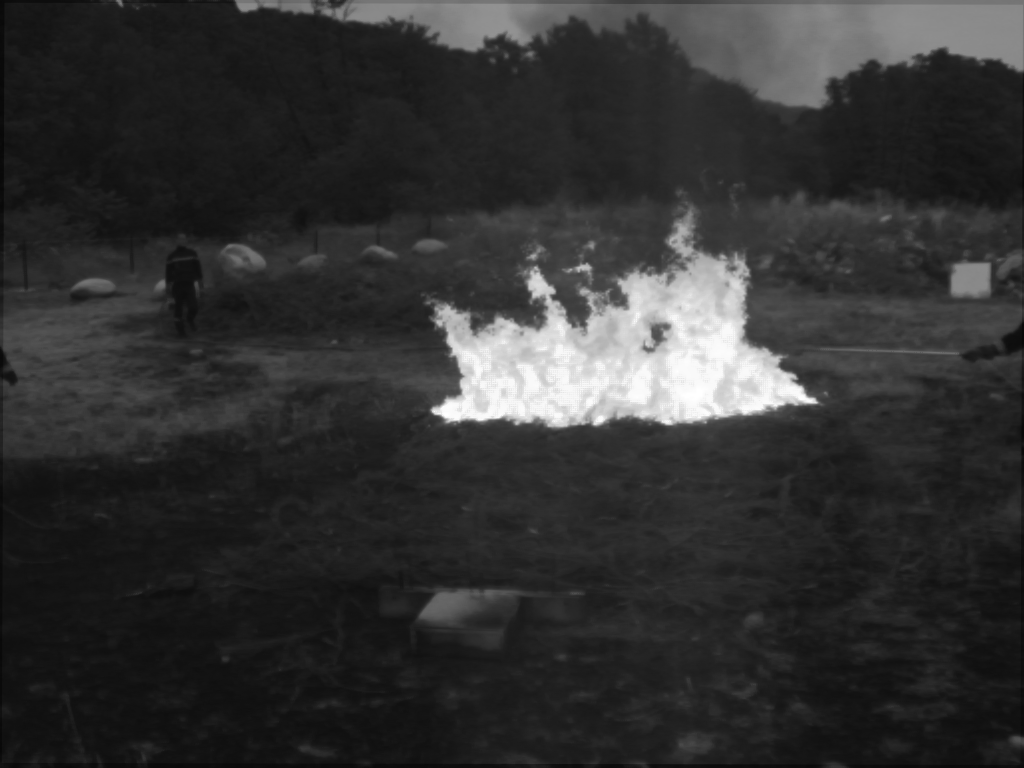}
  \caption{}
  \label{fig:vgg19_three_methods}
\end{subfigure}\hfil 
\begin{subfigure}{0.19\textwidth}
  \includegraphics[width=\linewidth]{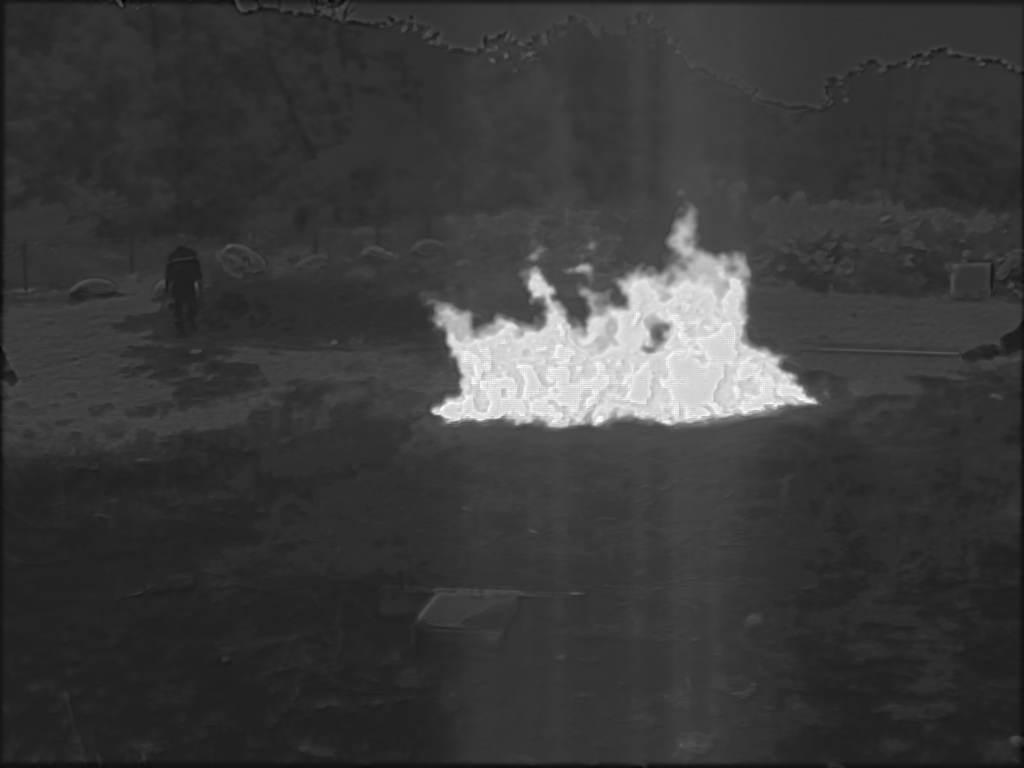}
  \caption{}
  \label{fig:fusiongan_three_methods}
\end{subfigure}\hfil 
\begin{subfigure}{0.19\textwidth}
  \includegraphics[width=\linewidth]{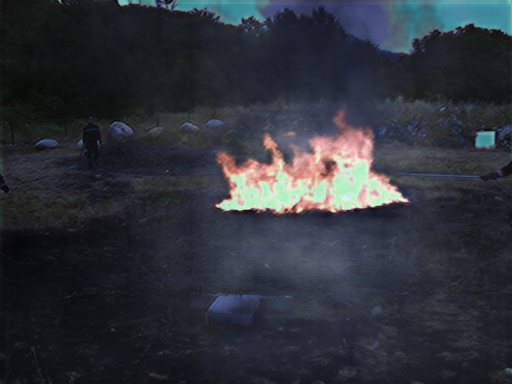}
  \caption{}
  \label{fig:firegan_three_methods}
\end{subfigure}

\caption{Sample resulting images from the three methods. In column ~\ref{fig:rgb_three_methods} are the RGB images, in ~\ref{fig:ir_three_methods} the IR images, in ~\ref{fig:vgg19_three_methods} the fused images from the \emph{VGG19} method, in ~\ref{fig:fusiongan_three_methods} the fused images from the \emph{FusionGAN} method, and in ~\ref{fig:firegan_three_methods} the fused images from the \emph{FIRe-GAN} method.}
\label{fig:three_methods_imgs}
\end{figure*}

Due to the improvement displayed by the U-Net architecture for G1 on the proposed ~\emph{FIRe-GAN}, we took this extended model and used it for its comparison with the works by Li et al. ~\cite{Li18} and Ma et al. ~\cite{Ma19_GAN}. For consistency and to make the comparison fair with these methods, we pre-trained our proposed extended model with the RGB-NIR dataset. Then, we tested the three models on the Corsican Fire Database. In this way, we were able to assess the generalization capabilities of these models on the new domain of fire imagery. Fig. ~\ref{fig:three_methods_boxplots} displays the results, Table ~\ref{tab:full_avg_results} shows the average results on the first three columns, and Fig. ~\ref{fig:three_methods_imgs} shows sample images produces by the three methods. We can observe that the \emph{VGG19} method presents the most balanced inclusion of information and, on average, higher quantitative results. Of the GAN-based models, the \emph{FusionGAN} one heavily favors the inclusion of thermal data, while the \emph{FIRe-GAN} model shows balanced results regarding the inclusion of source information; however, the metric results are lower on average than those of the \emph{VGG19} method.

\subsection{Transfer learning}
\label{transfer_learning}

The \emph{VGG19} and \emph{FusionGAN} methods can rely on source infrared images to perform the fusion process. In contrast, the ~\emph{UnmatchGAN} model and, by extension, the proposed ~\emph{FIRe-GAN} model, must generate approximate infrared ones and fuse them with the source visible ones. Then, it stands to reason that it could be more problematic for this model to generalize to new domains. As such, we perform a transfer learning phase with a segment of the fire images of the Corsican Fire Database and evaluate its performance.

We segment the dataset into a train and a validation sets. The training set has 8192 images after data augmentation, and the validation set consists of 128 image pairs. We set the training epochs to 3. Due to the strong thermal characteristics of fire images, we add a constant term $\gamma$ that multiplies the element of the loss function of \emph{G2} that represents the adversarial loss between \emph{G2} and \emph{D1}, thus prioritizing the inclusion of visible information. The final result is a balanced inclusion of visible and infrared information for the fire fused images. Experimentally, we found the best value of $\gamma$ to be 4.5. All other training parameters are the same as those mentioned in section ~\ref{arch_modification}, and all other loss functions remain the same as those of the \emph{UnmatchGAN} model. The modified loss function for \emph{G2} is as follows:

\begin{equation} \label{eq1}
\begin{split}
L_{G_2} = \gamma[\frac{1}{N}\sum_{n=1}^{N}(D_1(I^n_F)-c_1)^2] + \frac{1}{N}\sum_{n=1}^{N}(D_2(I^n_F)-c_2)^2 + \\
 \frac{\lambda}{HW}(\| I_F - I_R\|_F^2 + \xi\| \nabla I_F - \nabla I_V\|_F^2)
\end{split}
\end{equation}

\begin{figure*}
     \centering
     \begin{subfigure}[b]{0.43\textwidth}
         \centering
         \includegraphics[width=\textwidth]{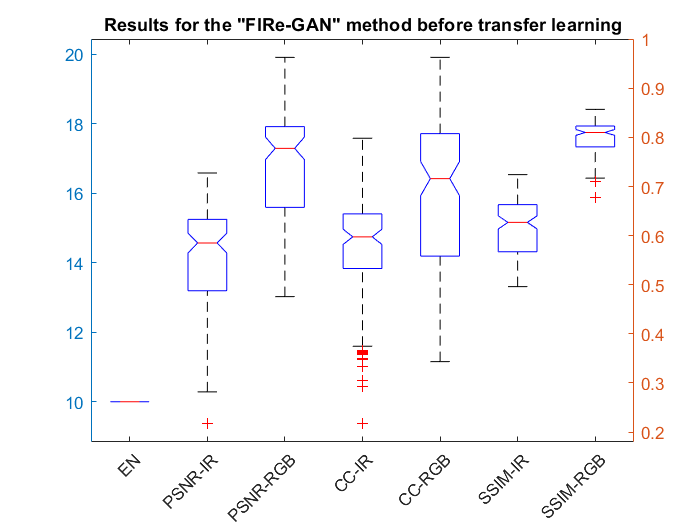}
         \caption{Metric results before transfer learning.}
         \label{fig:firegan_val_no_trans}
     \end{subfigure}
     \hfill
     \begin{subfigure}[b]{0.43\textwidth}
         \centering
         \includegraphics[width=\textwidth]{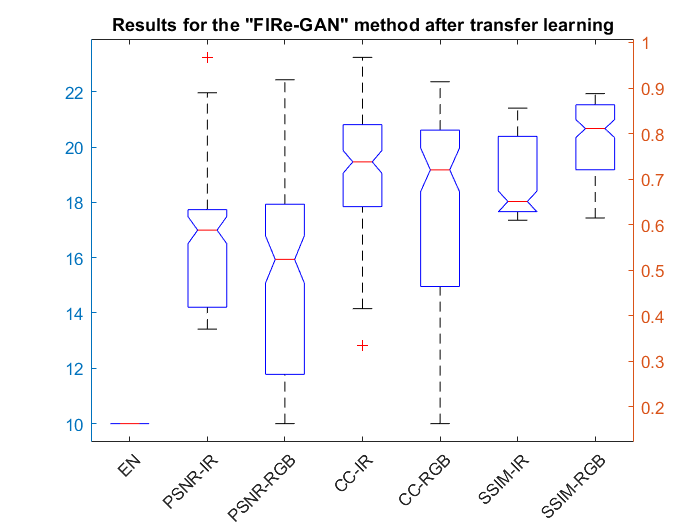}
         \caption{Metric results after transfer learning.}
         \label{fig:firegan_val_trans}
     \end{subfigure}
        
        \caption{Results for the ~\emph{FIRe-GAN} model from the fire images of the validation set before and after transfer learning.}
        \label{fig:transfer_learning_boxplots}
\end{figure*}

In Fig. ~\ref{fig:transfer_learning_boxplots} we show the results on the 128 images of the validation set before and after the transfer learning phase. Table ~\ref{tab:full_avg_results} shows the condensed average results on the last two columns. After only three training epochs, we can observe a marked improvement in the generated infrared images, as well as more accurate inclusion of thermal information on the fused ones.

\begin{figure*}[htb]
    \centering 

\begin{subfigure}{0.16\textwidth}
  \includegraphics[width=\linewidth]{505_rgb.png}
\end{subfigure}\hfil 
\begin{subfigure}{0.16\textwidth}
  \includegraphics[width=\linewidth]{505_nir.png}
\end{subfigure}\hfil 
\begin{subfigure}{0.16\textwidth}
  \includegraphics[width=\linewidth]{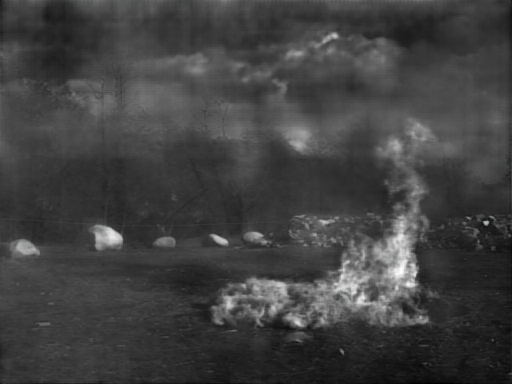}
\end{subfigure}\hfil 
\begin{subfigure}{0.16\textwidth}
  \includegraphics[width=\linewidth]{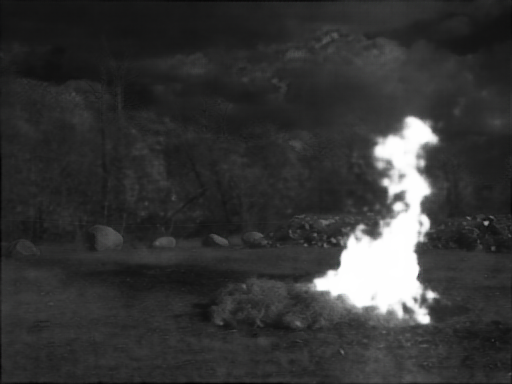}
\end{subfigure}\hfil 
\begin{subfigure}{0.16\textwidth}
  \includegraphics[width=\linewidth]{525fused_no_trans.png}
\end{subfigure}\hfil 
\begin{subfigure}{0.16\textwidth}
  \includegraphics[width=\linewidth]{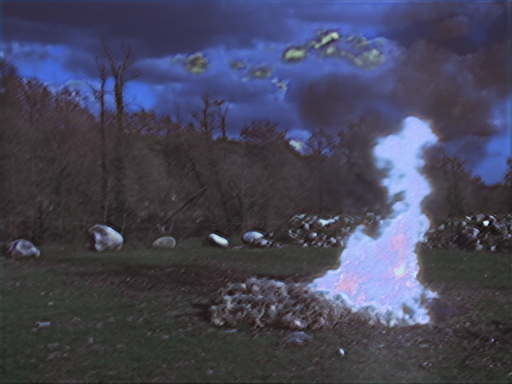}
\end{subfigure}\hfil 

\medskip
\begin{subfigure}{0.16\textwidth}
  \includegraphics[width=\linewidth]{595_rgb.png}
\end{subfigure}\hfil 
\begin{subfigure}{0.16\textwidth}
  \includegraphics[width=\linewidth]{595_nir.png}
\end{subfigure}\hfil 
\begin{subfigure}{0.16\textwidth}
  \includegraphics[width=\linewidth]{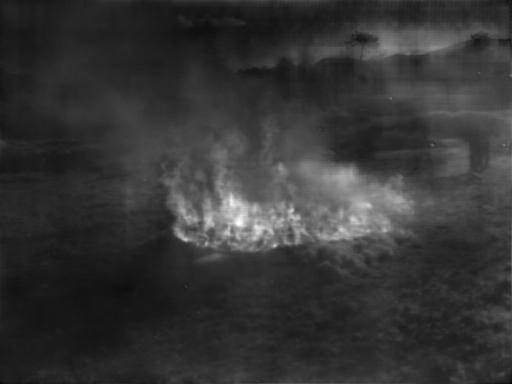}
\end{subfigure}\hfil 
\begin{subfigure}{0.16\textwidth}
  \includegraphics[width=\linewidth]{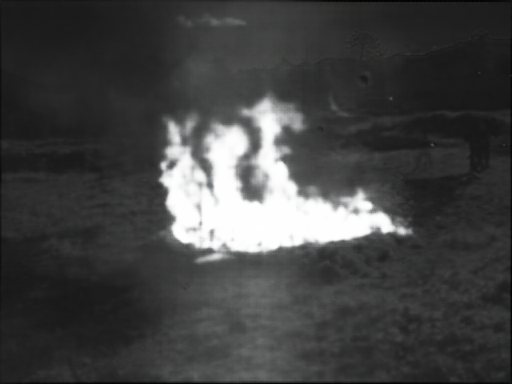}
\end{subfigure}\hfil 
\begin{subfigure}{0.16\textwidth}
  \includegraphics[width=\linewidth]{263fused_no_trans.png}
\end{subfigure}\hfil 
\begin{subfigure}{0.16\textwidth}
  \includegraphics[width=\linewidth]{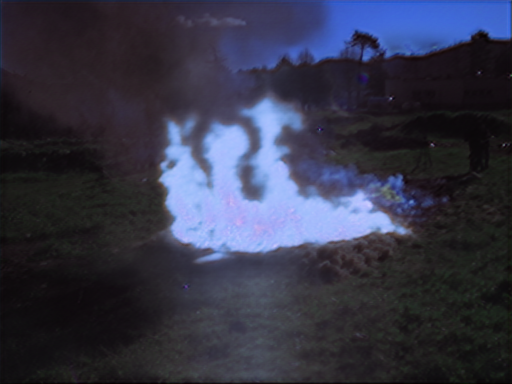}
\end{subfigure}\hfil 

\medskip
\begin{subfigure}{0.16\textwidth}
  \includegraphics[width=\linewidth]{078_rgb.png}
  \caption{}
  \label{fig:rgb_transfer}
\end{subfigure}\hfil 
\begin{subfigure}{0.16\textwidth}
  \includegraphics[width=\linewidth]{078_nir.png}
  \caption{}
  \label{fig:ir_transfer}
\end{subfigure}\hfil 
\begin{subfigure}{0.16\textwidth}
  \includegraphics[width=\linewidth]{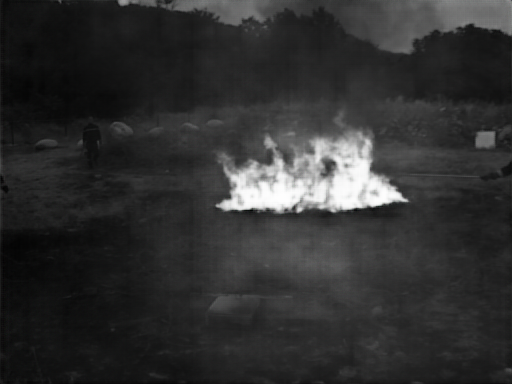}
  \caption{}
  \label{fig:gir_before_transfer}
\end{subfigure}\hfil 
\begin{subfigure}{0.16\textwidth}
  \includegraphics[width=\linewidth]{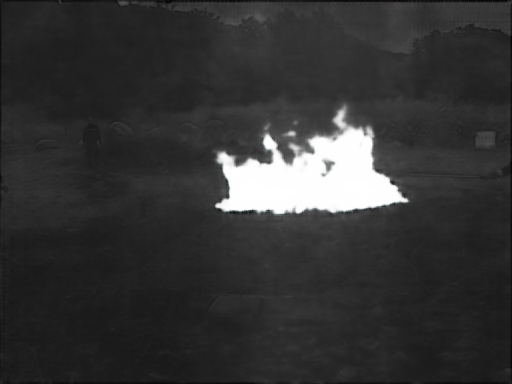}
  \caption{}
  \label{fig:gir_after_transfer}
\end{subfigure}
\begin{subfigure}{0.16\textwidth}
  \includegraphics[width=\linewidth]{211fused_no_trans.png}
  \caption{}
  \label{fig:fused_before_transfer}
\end{subfigure}
\begin{subfigure}{0.16\textwidth}
  \includegraphics[width=\linewidth]{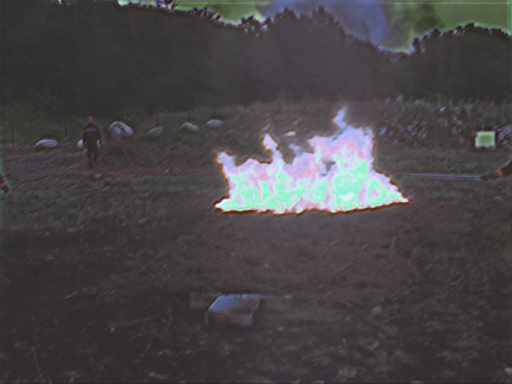}
  \caption{}
  \label{fig:fused_after_transfer}
\end{subfigure}

\caption{Sample resulting images before and after transfer learning. In column ~\ref{fig:rgb_transfer} are the RGB images, in ~\ref{fig:ir_transfer} the IR images, in ~\ref{fig:gir_before_transfer} the artificial IR images before transfer learning, in ~\ref{fig:gir_after_transfer} the artificial IR images after transfer learning, in ~\ref{fig:fused_before_transfer} the fused images before transfer learning, and in ~\ref{fig:fused_after_transfer} the fused images after transfer learning.}
\label{fig:transfer_learning_imgs}
\end{figure*}

\subsection{Summary}
\label{summary_results}

In Table ~\ref{tab:full_avg_results} we summarize the average results both for the evaluation of the three methods on the full Corsican Fire Database, as specified in section ~\ref{method_comp}, and of the transfer learning results on the validation set of the Corsican Fire Database as specified on section ~\ref{transfer_learning}.

\begin{table}
\caption{Average results for the three evaluated methods on the full Corsican Fire Database as specified on section ~\ref{method_comp} on the first three columns, and for the \emph{FIRe-GAN} method before and after transfer learning on the validation set of the Corsican Fire Database (section ~\ref{transfer_learning}) on the last two columns.}
\label{tab:full_avg_results}       
\begin{tabular}{llllll}
\hline\noalign{\smallskip}
Metric          & VGG19     & FusionGAN     & FIRe-GAN   & Before  & After   \\
\noalign{\smallskip}\hline\noalign{\smallskip}
EN              & 6.3415    & 6.0072        & 10         & 10      & 10      \\
CC IR-fused     & 0.7976    & 0.9657        & 0.5650     & 0.5774  & 0.7270  \\
CC RGB-fused    & 0.7637    & 0.4104        & 0.6651     & 0.6799  & 0.6499  \\
PSNR IR-fused   & 19.2014   & 22.5090       & 14.2065    & 14.3196 & 16.4212 \\
PSNR RGB-fused  & 19.2255   & 15.3900       & 16.7435    & 16.7687 & 15.6484 \\
SSIM IR-fused   & 0.8337    & 0.8389        & 0.6129     & 0.6187  & 0.7064  \\
SSIM RGB-fused  & 0.9007    & 0.7783        & 0.8000     & 0.8020  & 0.7935  \\
\noalign{\smallskip}\hline
\end{tabular}
\end{table}

\section{Discussion}
\label{discussion}

Of the three evaluated methods, the \emph{VGG19} one by Li et al. ~\cite{Li18} displayed the best overall performance for the new domain of fire imagery. This model is also the one that shows a more balanced behavior in terms of the inclusion of both visible and thermal information. The latter could be due to the way the authors leverage the VGG19 DL model. As Li et al. employ only the feature extraction capabilities of a pre-trained model, they do not need to train it for the particular task of image fusion. As the model was pre-trained on ImageNet, it demonstrates significant feature extraction capabilities, which explains the superior performance.

The ~\emph{FusionGAN} model proposed by Ma et al. ~\cite{Ma19_GAN} is relevant since it is the first attempt to use GANs for image fusion. The simplicity of an end-to-end model is also desirable. However, when applied to the new domain of fire images, this method tends to incorporate more thermal than visible information. This can be due to the fact that fire images have more well-defined thermal information, whereas non-fire images in the training set do not exhibit that strong of a distinction between thermal and visible images.

Our proposed ~\emph{FIRe-GAN} model has the advantages of being able to work with higher resolution images and to output three-channel fused images. This last feature allows it to learn to preserve color. Before performing transfer learning, it shows a balanced approach towards the inclusion of visible and thermal information; however, the overall performance is lower compared to the other two methods. Also, the generated IR images are very close to the source visible images; the model does not compensate for thermal information hidden behind features like smoke. Upon visual inspection, we can observe that the fused images preserve colors similar to the visible ones. 

When applying transfer learning to our proposed method on a segment of the Corsican Fire Database, after only three training epochs, the model can produce artificial IR images that are very close to the original ones, and fused images containing a balanced combination of thermal and visible information. This speaks well of the specialization capabilities of the model. It is also relevant to note that, even though the fused images preserve color information, the said color is no longer the same as the visible images, particularly on the fire regions. Since the color of the source visible and infrared images are significantly different for the case of fire images, it appears to be that the model learns to seek an intermediate color representation between the two.

Finally, it is worth remembering that we trained the model to generate approximate NIR images. The performance could change if the training set contains a different type of infrared image (SWIR, MWIR, or LWIR).

\section{Conclusions and future work}
\label{conclusions_future_work}

In the present paper, we demonstrate the feasibility of DL-based methods for the particular task of fire image fusion. The framework proposed by Li et al. ~\cite{Li18} displays the best overall performance. This method takes advantage of the feature extraction capabilities of DCNNs and traditional image fusion techniques for an effective combination.

The evaluated GAN-based methods show promise due to the simplicity of their implementation and their generalization and specialization capabilities. In particular, our proposed \emph{FIRe-GAN} model displays a balanced approach towards the inclusion of visible and infrared information, with consistent color preservation. Also, it is relevant to note that there is no much data available by DL standards, of visible-infrared image pairs, especially on the fire domain; the generation of more visible-infrared image pairs would improve the performance of the models and reduce the risk of overfitting. Finally, further experimentation is needed to assess the significance of color preservation on the fused images for different fire detection techniques.

\begin{acknowledgements}
The authors would like to thank the University of Corsica for providing access to the Corsican Fire Database.
\end{acknowledgements}

%
\section*{Conflict of interest}

The authors declare that they have no conflict of interest.

\bibliographystyle{spmpsci}      
\bibliography{references}   

%
%

\end{document}